# HEART DISEASE PREDICTION USING CASE BASED REASONING (CBR)

**MOHAIMINUL ISLAM BHUIYAN[1], CHAN HUE WAH[2], NUR SHAZWANI KAMARUDIN[3*], NUR HAFIEZA ISMAIL[4], AHMAD FAKHRI AB NASIR[5]**

[1,2,3,4,5]Faculty of Computing, University Malaysia Pahang Al-Sultan Abdullah

E-mail: nshazwani@umpsa.edu.my
*corresponding author

## ABSTRACT

This study provides an overview of heart disease prediction using an intelligent system. Predicting disease accurately is crucial in the medical field, but traditional methods relying solely on a doctor's experience often lack precision. To address this limitation, intelligent systems are applied as an alternative to traditional approaches. While various intelligent system methods exist, this study focuses on three: Fuzzy Logic, Neural Networks, and Case-Based Reasoning (CBR). A comparison of these techniques in terms of accuracy was conducted, and ultimately, Case-Based Reasoning (CBR) was selected for heart disease prediction. In the prediction phase, the heart disease dataset underwent data pre-processing to clean the data and data splitting to separate it into training and testing sets. The chosen intelligent system was then employed to predict heart disease outcomes based on the processed data. The experiment concluded with Case-Based Reasoning (CBR) achieving a notable accuracy rate of 97.95% in predicting heart disease. The findings also revealed that the probability of heart disease was 57.76% for males and 42.24% for females. Further analysis from related studies suggests that factors such as smoking and alcohol consumption are significant contributors to heart disease, particularly among males.

## 1. INTRODUCTION

Heart disease has caused a high level of concern among researchers since one of the most challenging aspects was to get an accurate and correct prediction [1]. The word "heart disease" describes various heart problems [2]. The four most frequent kinds of heart disease are coronary artery disease (CAD), arrhythmia, heart valve disease, and heart failure. According to a World Health Organization (WHO) study, many people worldwide suffer from heart disease yearly [3]. In the United States (U.S.), heart disease has become the top cause of mortality among people [4].

Various techniques and tools are used to predict heart disease, but it seems inefficient in the medical field [5]. This is because most methods are inadequate in calculating or predicting the outcome of heart disease in individuals, or the equipment is too expensive. Therefore, it was too challenging for them to predict heart disease. These challenges motivate us to research the prediction system which can predict heart disease accurately. Hence, an intelligent system technique will be applied to identify heart disease to resolve this concern correctly. An intelligent system (ML) is a part of artificial intelligence (AI) that helps researchers improve their prediction accuracy without being explicitly taught. Intelligent system algorithms use previous data as input to predict output results or outcomes [6]. These data's attributes or features will be utilized to identify the heart disease outcome, such as positive,1 or negative,0. Intelligent system algorithms are effective in predicting outcomes because they can handle massive amounts of data [7] and are capable of accurately predicting the development of heart disease [8]. As a result, in this study, intelligent system approaches will be studied and presented to predict the heart disease result correctly.

Heart disease has become a disease that can cause a person's death if not predicted accurately as there are many factors of heart disease. Coronary artery disease (CAD), arrhythmia, heart valve dysfunction, and heart failure are a few examples. Therefore, it was challenging to identify whether the patient had heart disease or not. In a traditional approach, the doctors in the hospital or clinic will base on their experience and knowledge to diagnose heart disease for every patient [1]. Therefore, the traditional approach will lead to inaccurate predictions. This is because sometimes there will be human errors like the doctors wrongly predicting the heart disease result based on their experience. The hospital and clinic also collect and keep many of their patient result records in a folder. Those result records will leave behind and become raw data. This method can lead to negative biases, errors, and additional medical costs, all of which influence the quality of care delivered to patients. Heart disease can be predicted using modern ways such as by using intelligent systems. By using intelligent system methods, the doctors only require keying in the data according to each input variable to get the outcome. These approaches will aid the doctors in hospital or clinic to provide the outcome result to the patients, but it has limitations. Many intelligent system techniques can be used for prediction, but only some methods are suitable. This is because some intelligent system techniques will only produce the images or text outcome where the outcome for the heart disease prediction is in binary form. Besides that, those intelligent system techniques that are suitable to predict the result of heart disease have a critical problem which is accuracy. Every intelligent system model uses different algorithms to predict the heart disease result. The data cleansing procedure is very significant in deciding the accuracy of the results. The practice of fixing or eliminating erroneous, corrupt, poorly formatted, duplicate, or incomplete data from a dataset is known as data cleaning. It is essential to identify and remove errors and duplicate data to establish a dependable dataset and increase the accuracy of the final result. Therefore, each intelligent system model will have different accuracy in predicting the resulting outcome.

The goal of this research is to develop an advanced intelligent system designed to enhance the accuracy of heart disease prediction, addressing existing limitations in conventional diagnostic methods.

## 2. RELATED WORKS

The authors of this research paper [9], Tanmay Kasbe & Ravi Singh Pippal used the Fuzzy Logic Intelligent system to diagnose heart disease. Fuzzy logic is a powerful reasoning method that best deals with uncertain data. Heart disease diagnosis is an essential criterion in daily life. Yet due to a lot of uncertainty and risk factors, heart disease diagnosis can sometimes be difficult for experts. When a heart attack is detected, the speed of detection is critical to save the patient's life and prevent heart damage.

The authors used the heart disease dataset from the UCI(Intelligent System Repository) to research. The dataset includes ten input variables: Systolic Blood Pressure, Serum Cholesterol, Maximum Heart Rate, Chest Pain, Fasting Blood Sugar, Old Peak, Electrocardiography (ECG), Thallium Scan, gender, and age. The output attribute will be the Result. Next, the tool used by the authors to diagnose the heart disease result was MATLAB. The accuracy of the study article, heart disease diagnostic system using Fuzzy Logic, is 93.33%. Fuzzy Logic Intelligent system diagnoses heart diseases through a fuzzy expert system with membership function, fuzzy input and output variables, and a fuzzy rule base.

In the first stage, fuzzy membership functions will be used for implementation in the MATLAB tool. The fuzzy membership function converts the crisp input from the heart disease dataset to provide the fuzzy inference system. The authors selected two membership functions such as triangular function and trapezoidal function. The lower limit, 'a' an upper limit, 'b' and a value of 'm' will clarify the triangular function. The range of the 'm' value will be a < m < b. Equation (1) below shows the triangular function.

$$\mu_A(x) = \begin{cases} 0, & x \le a \\ \dfrac{x-a}{m-a}, & a < x \le m \\ \dfrac{b-x}{b-m}, & m < x < b \\ 0, & x \le b \end{cases} \quad (1)$$

Then in the trapezoidal function, 'a' represents the lower limit, and 'd' represents the upper limit. Besides that, it also has a lower support limit defined as 'b' and an upper Support limit 'c'. Equation (2) shows the trapezoidal function.

$$\mu_A(x) = \begin{cases} 0, & (x < a) \, or \, (x > d) \\ \dfrac{x-a}{b-a}, & a \le x \le b \\ 1, & b \le x \le c \\ \dfrac{d-x}{d-c}, & c \le x \le d \end{cases} \quad (2)$$

The fuzzy expert system is vital in the second step for determining the input and output variables of the heart disease dataset. The UCI (Intelligent System Repository) heart disease dataset contains 10 input variables and 1 output variable. The fuzzy data rule base will be declared at the final stage. The fuzzy rule base, one of the most critical aspects of the fuzzy inference system, determines the quality of the heart disease outcome. The fuzzy rule is a conditional statement. IF-THEN statements give the form of fuzzy rules. The fuzzy data rule base is declared using AND/OR logic operator with either a single attribute or a combination of attributes. The authors reported 86 rules by using the right combination of attributes. Thus, the more the new rule base, the higher the accuracy. The advantage of Fuzzy Logic is it is dynamic and allows for rule changes. It even accepts imprecise input information. Then the disadvantage of this approach is the accuracy of these systems is compromised since they rely on inaccurate data and inputs. However, the limitation of this technique is that the rule of fuzzy logic is based on predefined rules, and if the rules are flawed, the result is predicted.

Heart disease was predicted by using the Neural Network Intelligent system technique or algorithm [10]. According to the researchers, developing a medical diagnosis system based on machine learning for heart disease prediction provides a more accurate diagnosis and lowers treatment costs. For the prediction system, the Backpropagation Algorithm, a widely used Artificial Neural Network learning methodology, was used to satisfy this need. The dataset of this research was taken from the Cleveland database. This dataset has 303 cases and 76 attributes, but the authors only used 14. Table 1 below shows all the attributes of the heart disease dataset in table form. In this study, 13 variables will be used as input, and 1 attribute, Num will be used as output. The authors of this study applied MATLAB R2015a tools to identify heart disease using a neural network approach. Backpropagation Algorithm with Artificial Neural Network (ANN) learning approach will be utilized in this study to predict heart disease.

*Table 1: Cleveland Heart Disease Dataset Attributes*

| Clinical Features | Description |
|---|---|
| Age | Age |
| Ca | Number of major vessels (0-3) colored by fluoroscopy |
| Chol(mg/dl) | Serum cholesterol |
| Cp | Chest pain type |
| Exang | Exercise induced angina |
| Fbs | Fasting blood sugar |
| Num | Diagnosis of heart disease |
| Oldpeak | ST depression induced by exercise relative to rest |
| Restecg | Resting electrocardiographic results |
| Sex | Gender |
| Slope | The slope of the peak exercise ST segment |
| Thal | 3 = normal; 6 = fixed defect; 7 = reversible defect |
| Thalach | Maximum heart rate achieved |
| Trestbps(mmHg) | Resting Blood Pressure |

A multilayer perceptron neural network was built using the Artificial Neural Network (ANN) learning approach for the heart disease prediction system [11]. It contains three layers: an input layer, a concealed layer, and an output layer. The input layer will have 13 neurons because the dataset has 13 input variables. Next, the authors used 3 neurons for the hidden layer. The number of neurons in the input layer will be increased by one at a time by measuring their performance and selecting the best one. It will be good if the number of hidden layer neurons is equal or the same as the neuron of the input and output layers. Then the output layer will have 2 neurons. This is because the value of the output in the dataset was either disease absence or disease presence represented by 0 and 1 respectively.

Besides that, the Backpropagation Neural Network algorithm (BA) will be used to build multilayer neural networks. It is also known as an error-back propagation algorithm since it uses an error-correction learning rule base. The heart disease dataset will be split into 3 parts which are training, testing, and validation. The training data will be used in the process of Backpropagation. At first, a small random number will be used as the weightage for all the networks. The training data is then used as input, and the output for each unit is computed using the Sigmoid Function,

$$o = \sigma(\vec{W}.\vec{x}), \sigma(y) = \frac{1}{1+e^{-y}} \qquad (3)$$

where $\vec{W}$ is the vector unit of weightage values while $\vec{x}$ is the vector unit for network input values. After that, error calculation will begin with the error signal ($\delta$) calculation for every network output that propagated as input to all neurons in the network. The first error term, $\delta_k$ with the equation of

$$\delta_k \leftarrow o_k(1-o_k)(t_k - o_k) \qquad (4)$$

will be used to compute for every network output unit where $o_k$ reflect the network output for output unit $k$ and define the target output for output unit $k$. The second error term, $\delta_h$ with the equation of

$$\delta_h \leftarrow o_h(1-o_h) \sum_{k \, \epsilon outputs} W_{kh} \delta_k \qquad (5)$$

will be used to compute for every hidden unit $h$ where $W_{kh}$ show the network weightage from the hidden unit $h$ to the output unit $k$. Therefore, every network weightage will be updated by using the equation of

$$W_{ji} \leftarrow W_{ji} + \Delta W_{ji} \qquad (6)$$

Where,

$$\Delta W_{ji} = \eta \delta_j x_{ji} \qquad (7)$$

The $\eta$ indicate the learning rate while $x_{ji}$ refers to the input from unit $i$ and unit $j$. The accuracy result from this study was 95%.

A decision support system in heart disease diagnosis by case-based recommendation author is proposed by Prinsha Prakash [12]. According to the authors, most medical decisions must be made quickly, simply relying on the Doctor's unaided memory. Continuous training and recertification procedures push the Doctor to retain more important information in mind at all times, but limitations of human memory and recall, along with the expansion of knowledge, ensure that most of what is known cannot be understood by most people. To identify the cardiac condition that was the result of this research, the author used a method known as Case-Based Reasoning (CBR), which is an intelligent system. These methods can assist in organizing, storing, and retrieving the information that is necessary when dealing with each problematic case, as well as propose appropriate diagnostic, prognostic, and therapeutic judgments and decision-making processes.

Case-based reasoning differs in several ways from other AI strategies like knowledge-based systems (KBS). CBR uses the particular knowledge of previously encountered, concrete issue situations rather than relying only on general knowledge of a problem area or drawing links along generalized relationships between problem descriptors and conclusions. Additionally, CBR provides incremental, continuous learning since new knowledge is maintained each time a problem is resolved and may be used to tackle similar issues in the future.

These authors use the heart disease dataset found at Cleveland Clinic Foundation. Table 1 provides a visual representation of the Cleveland Clinic Foundation's statistics on heart disease. There are fourteen properties in the dataset, thirteen of which are input variables and one of which is an output or result variable. The author used the Case-Based Reasoning (CBR) approach to diagnose cardiac illness resulting from this research. This was done with the help of the MATLAB program. The CBR method or algorithm is broken down into four stages: retrieved, reuse, revise, and retain.

Since the revise and retain process may be done manually, the author used the retrieve and reuse technique in this research. The author uses scanned images of 2D echocardiographic, EEG, ECG, and heart images as input, and image processing techniques are used to validate the normal or abnormal condition of the heart. Then the result data will be saved in the case base. The doctor will reuse this result data during the patient consultation by retrieving comparable instances from the case base and instantly providing the heart disease result. Aside from that, if a special case does not exist in the case base, the revise and retain approach will be utilized to get the result and preserve it in the case base for future usage. The advantage of Case-Based Recommendation (CBR) is that it is not complex to implement. CBR uses case-based knowledge, therefore the output or result can be proposed rapidly. Then the disadvantage of this approach is the CBR requires more previous data cases to predict the outcome accurately. So, it needs a huge amount of storage to store the previous cases. However, the limitation of this technique is It takes longer to compute the similarity between the training and testing data during the revision phases.

Accuracy is crucial in medical diagnosis systems, especially for heart disease where there is a need for more precise predictions. This research aims to develop an intelligent system capable of delivering more accurate heart disease predictions, surpassing the performance of previous systems.

## 3. METHODOLOGY

Since using the traditional approaches of the doctor's experience and knowledge to predict the heart disease result was complex and not accurate, intelligent system algorithms or techniques will be proposed to overcome the problem. Intelligent system algorithms or techniques can assist doctors in predicting heart disease results automatically by applying the formula to the intelligent system algorithm. Many different types of intelligent system algorithms are available today. Not all of them are suitable for predicting the heart disease outcome since the outcome is a binary number that can be either 0 or 1. This is because many intelligent system algorithms can only provide the outcome in images or text form [13]. Consequently, the three best intelligent system algorithms suited for predicting heart disease results are chosen, and only the best and most accurate algorithm is selected to predict the outcome.

The heart disease dataset is initially downloaded from Kaggle, followed by its pre-processing to enhance data quality. This involves applying data normalization techniques, ensuring uniformity in the data range. Subsequently, the dataset is divided into a 60:40 ratio for training and testing purposes. The Case-Based Reasoning (CBR) algorithm is then employed for heart disease prediction, using 60% of the data for training and 40% for testing. The accuracy of the predictions is determined by comparing them against the original dataset. The research framework and the detailed process flow are illustrated in Figures 1 and 2, respectively.

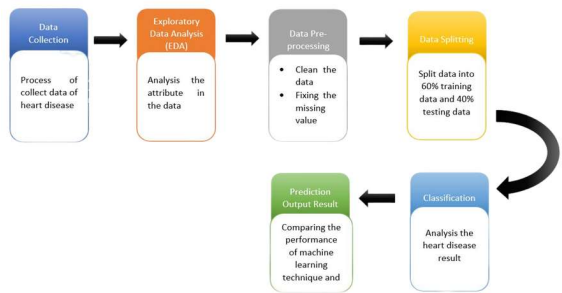

*Figure 1: Research framework*

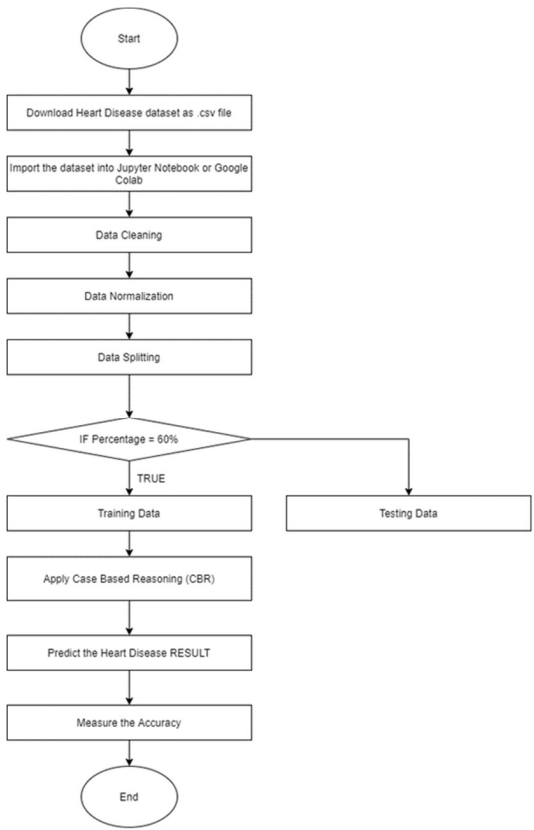

*Figure 2: Process flow chart*

The heart disease dataset, authored by David Lapp and sourced from Kaggle, comprises 1025 entries with 13 attributes and 1 output. Attributes include age, gender, chest pain (Cp), resting blood pressure (trestbps), serum cholesterol (Chol), fasting blood sugar (fbs), resting electrocardiographic results (Restecg), maximum heart rate achieved (thalach), exercise-induced angina (exang), ST depression induced by exercise relative to rest (Oldpeak), slope of the peak exercise ST segment (Slope), number of major vessels (Ca), and a blood disorder called thalassemia (thal). The dataset's characteristics, descriptions, and data types are detailed in figure 3.

| No | Attributes | Description | Datatypes |
|---|---|---|---|
| 1 | age | This indicates the person's age in years. | Integer |
| 2 | gender | This indicates the person's sex where:<br>Value 0 = Female<br>Value 1 = Male | Integer |
| 3 | cp | This indicates chest pain type:<br>Value 0 = Typical Angina<br>Value 1 = Atypical Angina<br>Value 2 = Non-anginal Pain<br>Value 3 = Asymptomatic | Integer |
| 4 | trestbps | This indicates the person's resting blood pressure (mmHg). | Integer |
| 5 | chol | This indicates the person's cholesterol measurements in mg/dl | Integer |
| 6 | fbs | This indicates the person's fasting blood sugar in measurement of less than 120 mg/dl where:<br>Value 0 = not less than 120 mg/dl<br>Value 1 = less than 120 mg/dl | Integer |
| 7 | restecg | This indicates resting electrocardiographic results. There are 3 values:<br>Value 0 = Showing definite left ventricular hypertrophy by Estes' criteria<br>Value 1 = Normal<br>Value 2 = Having ST-T wave abnormality | Integer |
| 8 | thalach | This indicates the person's maximum heart rate achieved. | Integer |
| 9 | exang | This indicates the exercise induced angina where:<br>Value 0 = no<br>Value 1 = Yes | Integer |
| 10 | oldpeak | This indicates the ST depression induced by exercise relative to rest. | Float |
| 11 | slope | This indicates the slope of the peak exercise ST segment.<br>Value 0 = downsloping<br>Value 1 = flat<br>Value 2 = upsloping | Integer |
| 12 | ca | The number of major vessels (0-3). | Integer |
| 13 | thal | This indicates where:<br>Value 1 = normal<br>Value 2 = fixed defect<br>Value 3 = reversable defect | Integer |
| 14 | target | It indicates the presence of the heart disease in the patients.<br>Value 0 = no heart disease<br>Value 1 = has heart disease | Integer |

*Figure 3: Attributes and Description of the Heart Disease Dataset*

Data splitting involves dividing the data into two or more parts. When there is a split into two parts, one is often utilized to analyze or test the data, while the other is used to train the model [14]. The process of data splitting is a crucial component of data science, especially for developing models based on data. Accuracy in the development of data models and the processes that apply data models, such as machine learning, may be improved with the aid of this approach. There are 3 types of data splitting. First is random sampling. The data modeling process is protected from bias using this data sampling strategy, which prevents bias toward various potential data features. However, there may be problems with the random splitting method due to the unequal distribution of the data. Secondly, stratified random sampling. This method picks random samples of data from a set of parameters. It ensures that the data in the training set and the test set are split up correctly. Thirdly, non-random sampling. Data modelers use this method to use the most current data as the test set.

In this research, the data will be split into the ratio 60:40, where 60% will be the training data and 40% will be the testing data using the non-random sampling method. This research requires the predicted dataset to compare with the previous dataset to identify the accuracy. In the heart.csv dataset, there are 1025 data. Therefore, 615 data will be used for training and 410 data will be used for testing.

Data preprocessing is the process of cleaning or dropping the data. It is an important step in the data mining process. Data pre-processing is to convert or transform the raw data into high-quality data or understandable format [15]. Data normalization is also part of data pre-processing. Normalization is a procedure that is often used in the process of data preparation or data pre-processing in machine learning. The process of transforming the different columns of a dataset to the same scale is referred to as normalization. The data in the dataset will be converted to a scale of 0 to 1 by using the normalization method. It is not necessary to normalize each dataset when using machine learning. It is only required in situations in which the ranges of the characteristics differ. The purpose of normalizing the data is to enhance the accuracy of the prediction.

The data that has been split into the ratio of 60:40 and normalized will be used in the Case-Based Reasoning (CBR) algorithm for prediction purposes. Case-based reasoning, often known as CBR, is an approach for resolving novel issues by modifying approaches that have already shown success in solving problems based on the past. It is an Artificial Intelligence (AI) technique that imitates how humans make a decision. Memory, learning, as well as planning, and problem-solving, are all investigated during CBR. This sets the way for a new technology, including intelligent computer systems that can solve issues and adapt to new circumstances. In case-based reasoning (CBR), the "intelligent" reuse of information from previously solved problems, commonly known as "cases," is predicated on the assumption that if two problems are similar, their solutions will also be similar. Cases may be thought of as examples of previously resolved issues. In the CBR algorithm, 4 stages will pass through: retrieve, reuse, revise, and retain shown in Figure 4.

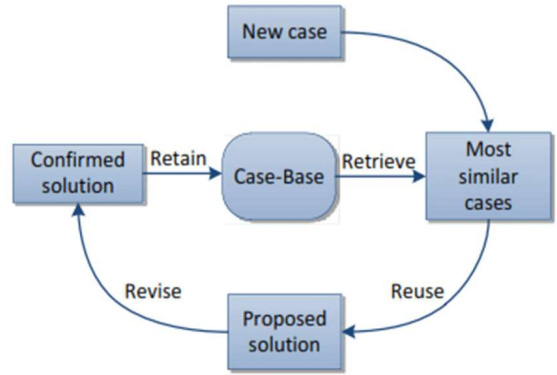

*Figure 4: Case based reasoning*

The first stage is retrieving. The similarities will be calculated between the testing data and training data. It is used in problem-solving and reasoning to match a previous experience/case (case base) with the new unseen problem to find a solution. It also can be called case matching in CBR matching new cases with the previous cases from the case base to find a solution. The purpose of the similarities is to select cases that can be adapted easily to the current problem and select cases that have (nearly) the same solution as the current problem. The basic assumption for the retrieved phase is "similar problems have similar solutions" and the goal of similarity modeling is to provide a good approximation. Two types of similarities need to be calculated such as local similarity and Global Similarity. Local Similarity is used to compute the similarity between query (new problem) and case attribute values while Global Similarity is a build-up from the number of local similarities function It is a weighted sum of the local similarity. The formulas for Local Similarity and Global Similarity are shown in Figure 5 and Figure 6 below.

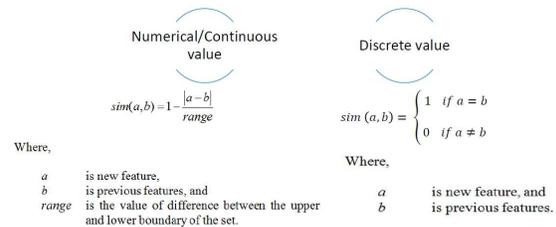

*Figure 5: Local Similarity Formula*

$$sim(A, B) = \frac{1}{\sum w_i} \cdot \sum_{i=1}^{p} w_i \cdot sim_i(a, b)$$

Where,

| | |
|---|---|
| $A$ | is new case, |
| $B$ | is previous cases, |
| $a$ | is new feature from local similarity, |
| $b$ | is previous features from local similarity, |
| $p$ | is the number of attributes, |
| $i$ | is the iteration |
| $w_i$ | is weight of attributes $i$ $\sum_{i=1}^{p} w_i = 1$, and |
| $sim_i$ | is local similarity calculate for attribute $i$. |

*Figure 6: Global Similarity Formula*

Thus, similarity measurement for local similarity is calculated between each attribute value, while Global Similarities are computed between each case. The steps in the retrieved phase are:

The data split into training and testing will be imported in the first step. It will import the original data and the data that had been normalized. Next, a variable will be used to declare for each data that is being imported. Then the output column of the normalized training data will be eliminated for prediction purposes.

In the next step, to calculate the local similarity for each case, the minimum and maximum data will be identified from the normalized training data and the range will be calculated by using the maximum value and subtracting the minimum value for each attribute. Besides, for global similarity calculation purposes, the weightage will be declared to be 1 for each attribute.

After that, the local similarity and global similarity algorithms will be applied to predict the outcome result of the heart disease data.

Next, after each testing data's local and global similarity is calculated, it will move to the reuse phase. The reuse process in the CBR cycle is responsible for proposing a solution for a new problem from the solutions in the retrieved cases where it reuses previous experiences as the solution to a new problem or situation. Reusing a retrieved case can be as easy as returning the retrieved solution, unchanged, as the proposed solution for the latest issue. It did not require any modification where it just copied back the solution in the previous problem that achieved the highest similarity as the new case or problem solution. The highest similarity among the global similarity result will be identified in the reuse phase. After the highest similarity has been identified, the result of the training case that achieves the highest global similarity will be appended as the result for the testing case data.

Besides, the revised phase determines how the solution can be used in the new setting. It aims to verify and ensure the correctness of the solution. Since the solution of the Heart Disease Data has only 2 values which are 0 and 1, then the revised cycle is not required.

The last cycle is the retained phase. The result for the new solution or test data will be merged and stored in the train data before normalizing. So, it will store and merge the test data with the new result to the train data before normalizing and saving it into the case based. Therefore, the new solution will be saved.

In this research, there are some constraints. This study only focuses on one intelligent system algorithm. Because many other techniques are not considered or offered in this study, claiming the best technique from other intelligent system techniques becomes the constraint. Next, since intelligent system approaches differ, there is only a limited amount of time to assess and research them all. Therefore, it is challenging to investigate, analyze, and evaluate all the other intelligent system approaches. The limitation of this heart disease prediction research is static. This heart prediction is not a real-time prediction. This is because the dataset taken from the website does not contain the actual latest heart disease data. Therefore, different heart disease datasets will have different results and accuracy.

4. **RESULTS**

Upon completion of the heart disease prediction and subsequent saving of the results, a comparative analysis will be conducted between the outputs derived from the original dataset and the normalized dataset. Both CSV files containing the prediction results will be imported into the system for evaluation. This comparison will focus on analyzing discrepancies or improvements in predictive performance between the two datasets, allowing for a detailed assessment of the impact of data normalization on the accuracy and consistency of the predictions.

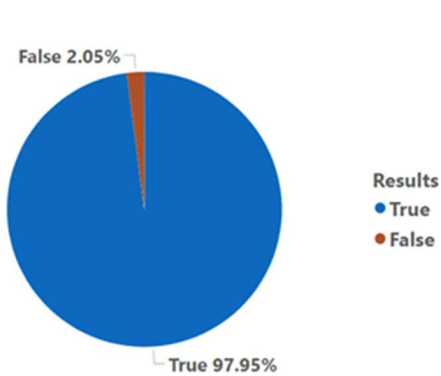

*Figure 7: Accuracy in Pie Chart*

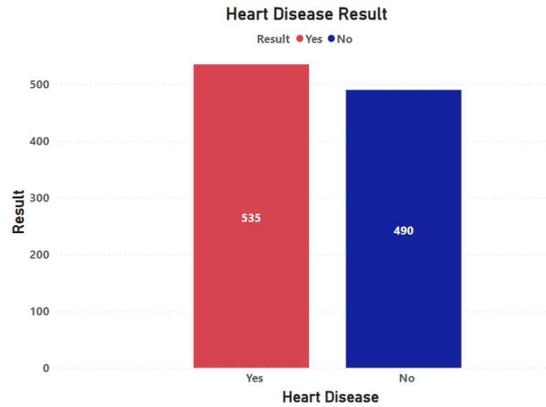

*Figure 9: Number of Data According to heart disease*

In our study, the Case-Based Reasoning (CBR) methodology yielded an impressive accuracy of 97.95% in predicting heart disease, as illustrated in Figure 7. This high level of precision highlights the effectiveness of CBR as a dependable tool in the medical diagnostic process, particularly for the early detection and treatment planning of heart disease. The results signify a notable advancement in the use of intelligent systems within healthcare, showcasing CBR's potential to improve diagnostic accuracy over traditional approaches.

From the result, a few visualization techniques will be applied to the predicted heart disease dataset and the result data using Microsoft Power BI. Figure 6 above shows the number of heart disease data according to gender. The results show there are 1025 data altogether and 713 data come from Males, and 312 from Females. From Figure 9 above, the number of patients with heart disease is 535 out of 1025.

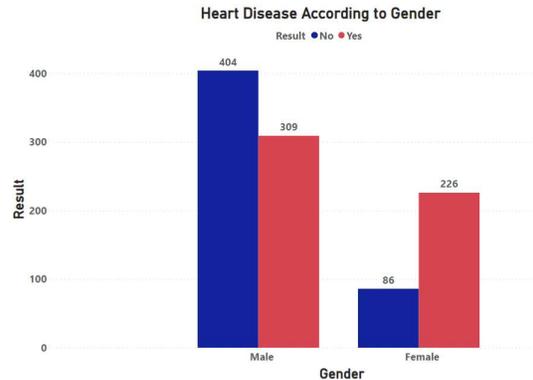

*Figure 10: Heart Disease According to Gender (Overall)*

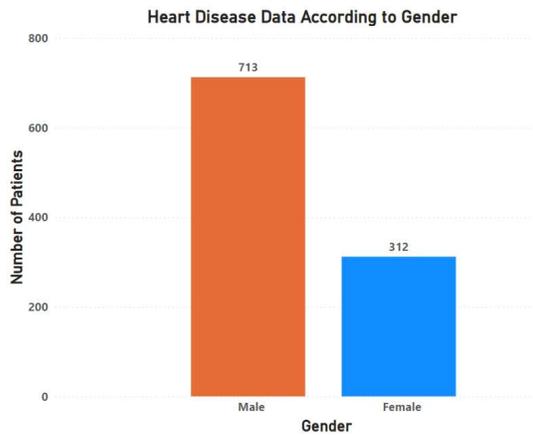

*Figure 8: Number of Data According to Gender*

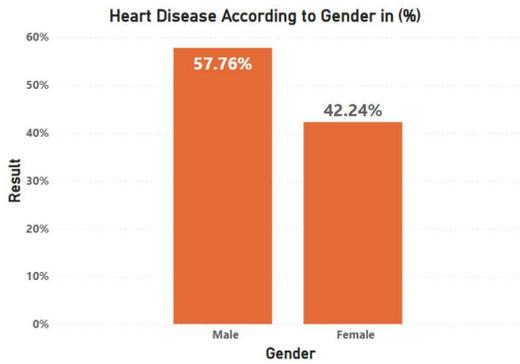

*Figure 11: Heart Disease According to Gender (in %)*

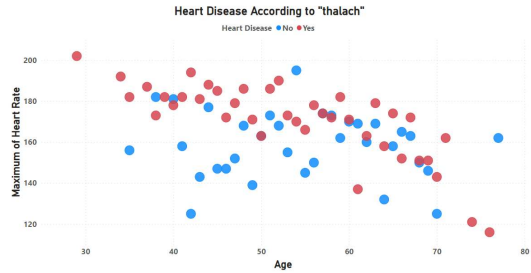

*Figure 13: Maximum of Heart Rate*

Figure 10 and 11 above shows the bar chart on heart disease according to gender that was visualized from the predicted result. The Figure 8 graph shows the overall number of patients who have and don't have heart disease according to gender. The Figure 9 graph result shows that the gender percentage of gender males getting heart disease is higher than that of females. The percentage for getting heart disease for males is 57.76% while for females, 42.24%. Based on the justification from [16] and [17] the factor activity that can cause heart disease, especially in males such as smoking and alcohol. Both activities are always done by a male more than compared to females.

Next, Figure 13 above shows the dotted graph visualized from the predicted result. The chart shows the Maximum Heart Rate according to age. The maximum heart rate is calculated from the attribute "thalach". It is the maximum heart rate that a person achieves. The result shows that the patient who is age 29 has the highest heart rate. According to [19], middle-aged people used to have a higher heart rate due to their daily activities. This is because, in that range of age, many people are stressed with their work and so on. Therefore, they are at high risk of getting heart disease if they do not have a healthy diet regularly.

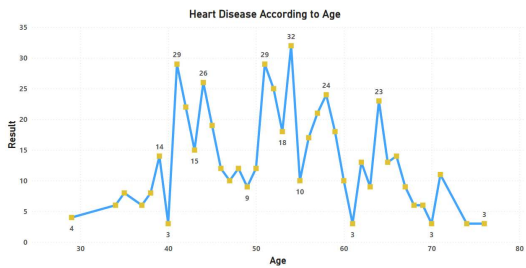

*Figure 12: Heart Disease According to Age*

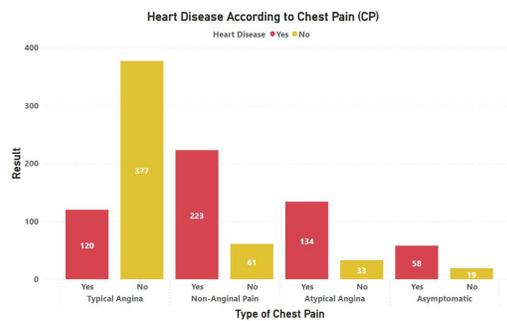

*Figure 14: Heart Disease According to Chest Pain*

Figure 12 above shows the line graph visualized from the predicted result. The chart shows the number of heart diseases according to age. The result shows that people aged 54 get the highest heart disease chance. There are 32 patients from the sample of 535 who get the highest heart disease at the age of 54. According to [18], most male patients focused on their work and didn't care about their health. So, it will lead to an unhealthy diet and slowly when the age grows older, they get a higher chance of having heart disease.

Besides that, Figure 14 shows the bar graph that is visualized from the predicted result. The graph shows the Chest Pain Type according to gender. The chest pain is calculated according to the gender from the attribute "cp". "cp" is a pain in any area of your chest. If you don't treat it right away, it could spread to other parts of your body, like your arms, neck, or jaw. Pain in the chest can feel sharp or dull. You might feel tight, achy, or like your chest is squeezed or crushed. Pain in the chest can last a few minutes or a few hours. It has four types of Chest Pain: Typical Angina, Atypical Angina, Non-anginal Pain, and Asymptomatic. Atypical pain is often defined as pain in the chest and abdomen or back, or as pain that is burning, stabbing, like stomach aches [20]. Pain in the chest, arm, or jaw that is dull, heavy, tight, or crushing is a common sign. Non-Anginal Pain is one of the chest pains caused by heart disease [21]. It feels like your chest is being squeezed or tightened or like there is pressure or weight on it, especially behind your sternum. We might feel it on the right, left, or right in the middle. Asymptomatic left ventricular systolic dysfunction (ALVSD), also called stage B heart failure, is a low left ventricular pulse rate function that doesn't cause any symptoms [22]. From the graph above, we can observe that the patients that get typical angina only have 120 patients that get the disease out of 497 patients. Next, for the non-anginal pain patients, 223 patients get the disease out of 284. Next, the Atypical Angina patients who get heart disease have 134 patients out of. Besides that, Asymptomatic patients with heart disease have 58 patients from 77.

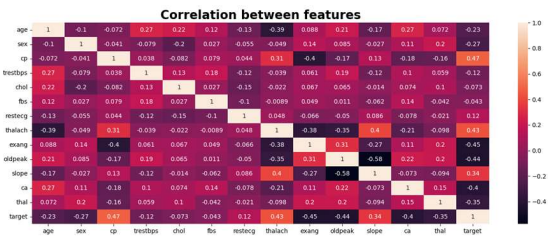

*Figure 15: Correlation Coefficient*

After that, the product-moment correlation coefficient technique may be used to determine whether there is a correlation between the attributes and the heatmap can then be constructed. The feature with the highest degree of correlation will be chosen for inclusion in the result of the model. The technique used to construct a predictive model is called feature selection [23]. The procedure has been used to limit the number of input variables by picking a relevant feature. In the field of machine learning, the reason we suggested using feature selection was so that we could improve the overall performance of the algorithm and get highly effective results when it came to training the machine. The technique of selecting the features that will most significantly contribute to the desired outcome of a prediction is known as "feature selection." This may be done either automatically or manually. Having unnecessary characteristics in the dataset will lead to a decline in the accuracy of the models; nevertheless, the machine will function more effectively if the unnecessary dataset is distinguished.

### 4.1 Comparison with prior works

Numerous approaches have been implemented in the past to predict heart disease, each yielding varying degrees of success. Despite these efforts, there is still a significant need for improvement, as the complexity and diversity of heart disease symptoms and risk factors make accurate prediction challenging. The advancements in machine learning (ML), deep learning (DL), and other intelligent systems have driven progress, but the potential for further enhancement remains substantial.

To highlight our contribution in advancing heart disease prediction, we present a comparative analysis (Table 2) between our work and previous studies. This comparison not only showcases the accuracy of different predictive models but also illustrates how our approach, utilizing Case-Based Reasoning (CBR), stands out in terms of performance.

In previous research, various methods such as Machine Learning, Deep Learning, Artificial Neural Networks (ANN), and Fuzzy Logic have been employed. For instance, Vayadande et al. [24] used a combination of ML and DL techniques, including Logistic Regression and XGBoost, achieving an accuracy of 88.52%. Pal et al. [25] applied Random Forest and obtained a slightly higher accuracy of 93.3%. Pasha et al. [26] employed an ANN approach, which resulted in an accuracy of 85.24%, while Kasbe et al. [9] and T.K. et al. [10] used Fuzzy Logic and ANN, reaching accuracies of 93.33% and 95%, respectively.

In contrast, our study leverages Case-Based Reasoning (CBR), a knowledge-based system that retrieves solutions based on the similarity of past cases to new ones. This method achieved a notable accuracy of 97.95%, which is higher than all the compared studies. The significant improvement in accuracy underscores the effectiveness of CBR in handling the complexity of heart disease prediction by learning from historical cases and adapting to new, unseen cases through its iterative process of retrieval and revision.

By providing this comparative analysis, we demonstrate how our research not only builds on but also advances previous work in heart disease prediction. The superior accuracy of our model suggests that CBR is a promising tool for enhancing the precision of diagnostic systems in the medical field, offering a more reliable method for early detection and intervention of heart disease.

*Table 2: Comparison with prior studies*

| Study | Method | Accuracy |
|---|---|---|
| Vayadande et al. [24] | ML and DL | 88.52% (Logistic regression, XGBoost) |
| Pal et al. [25] | Random Forest | 93.3% |
| Pasha et al. [26] | ANN | 85.24% |
| Kasbe et al. [9] | Fuzzy Logic | 93.33% |
| T.K. et al. [10] | ANN | 95% |
| **Ours** | **CBR** | **97.95%** |

## 5. CONCLUSION

This research successfully predicted heart disease using the intelligent system, Case-Based Reasoning (CBR). The heart disease prediction process involved the application of the CBR algorithm, which included data splitting and pre-processing steps such as data normalization. The CBR algorithm operates in four stages: Retrieve, Reuse, Revise, and Retain, with the Local and Global Similarity algorithm applied during the Retrieve stage. The dataset was divided into training and testing sets at a specific ratio, followed by data normalization to prepare it for the prediction phase.

In this study, the CBR algorithm was used to assess both local and global similarity within the training and testing data. The data exhibiting the highest similarity was selected as the predictive result for new cases. This approach proved effective, achieving a heart disease prediction accuracy of 97.95%.

However, some limitations were encountered during the research, particularly related to the dataset size and the nature of the CBR algorithm. The dataset comprised only 1,025 entries, which was insufficient for achieving optimal accuracy, as CBR performs best with larger datasets. Additionally, the CBR algorithm's process of comparing local and global similarities between each training instance is time-consuming. As a result, larger datasets significantly increase the time required for heart disease prediction, making the process more time-intensive as the dataset grows.

## ACKNOWLEDGEMENT

This work is supported by Faculty of Computing UMPSA and UMPSA Research Grant RDU230353.

## REFERENCES

[1] H. S. &. M. A. Rizvi, "Prediction of Heart Disease using Machine Learning Algorithms: A Survey," *International Journal on Recent and Innovation Trends in Computing and Communication,* pp. 99-104, 2017.

[2] S. Y. H. P. S. H. C. H. C. a. E. J. L. A. H. Chen, "HDPS: Heart disease prediction system," in *Computing in Cardiology*, Hangzhou, China, 2011.

[3] D. A. R. M. Ramalingam VV, "Heart disease prediction using machine learning techniques: a survey.," *International Journal of Engineering & Technology,* pp. 684-687, 2018.

[4] &. A. A.-M. Rahma Atallah, "Heart Disease Detection Using Machine Learning Majority Voting Ensemble Method," in *2nd International Conference on new Trends in Computing Sciences (ICTCS)*, Amman, Jordan, 2019.

[5] P. C. a. B. L. D. M. A. jabbar, "Prediction of risk score for heart disease using associative classification and hybrid feature subset selection," in *12th International Conference on Intelligent Systems Design and Applications (ISDA)*, Kochi, India, 2012.

[6] V. R. G. B. L. M. a. H. L. N. S. Kamarudin, "A Study of Reddit-User's Response to Rape," in *IEEE/ACM International Conference on Advances in Social Networks Analysis and Mining (ASONAM)*, Barcelona, Spain, 2018.

[7] A. A. ,. M. S. ,. D. R. D. P. G. Apurb Rajdhan, "Heart Disease Prediction using Machine Learning," *INTERNATIONAL JOURNAL OF ENGINEERING RESEARCH & TECHNOLOGY (IJERT) ,* 2020.

[8] D. P. S. &. B. S. Shah, "Heart Disease Prediction using Machine Learning Techniques," *SN Computer Science,* 2020.

[9] &. R. S. P. Tanmay Kasbe, "Design of heart disease diagnosis system using fuzzy logic," in *International Conference on Energy, Communication, Data Analytics and Soft Computing (ICECDS)*, Chennai, India, 2017.

[10] T. K. a. Ö. Kılıç, "Prediction of heart disease using neural network," in *International Conference on Computer Science and Engineering (UBMK)*, Antalya, Turkey, 2017.

[11] N. S. K. a. A. F. A. N. N. H. Ismail, "The Neuropsychology Assessment for Identifying Dementia in Parkinson's Disease Patients using a Deep Neural Network," in *International Conference on Software Engineering & Computer Systems and 4th International Conference on Computational Science and Information Management (ICSECS-ICOCSIM)*, Pekan, Malaysia, 2021.

[12] P. Prakash, "Decision Support System In Heart Disease Diagnosis By Case Based Recommendation," *International Journal of Scientific & Technology Research,* pp. 51-55, 2015.

[13] Kamarudin, N.S., Makhtar, M., Shamsuddin, S.N.W. and Fadzli, S.A. Shape-Based Single Object Classification Using Ensemble Method Classifiers. *International Journal on Advanced Science Engineering Information Technology*, 7(5), pp.1908-1912, 2017

[14] G. B. a. H. L. N. S. Kamarudin, "A Study on Mental Health Discussion through Reddit," in *International Conference on Software Engineering & Computer Systems and 4th International Conference on Computational Science and Information Management (ICSECS-ICOCSIM)*, Pekan, Malaysia, 2021.

[15] Mokhairi, M. and Engku Fadzli Hasan, S.A., Comparison of image classification techniques using CALTECH 101 dataset. *journal of theoretical and applied information technology*, 71(1), pp.79-86, 2015

[16] A. D. J. P. A. M. Hem C. Jha, "Plasma circulatory markers in male and female patients with coronary artery disease," *Heart and Lung: Journal of Acute and Critical Care,* pp. 296-303, 2010.

[17] M. G. L. A. B. D. L. Donald M Lloyd-Jones, "Lifetime risk of developing coronary heart disease," *The Lancet,* pp. 89-92, 1999.

[18] C. M. L. P. B. T. S. M. &. H. S. M. Alexander, "NCEP-defined metabolic syndrome, diabetes, and prevalence of coronary heart disease among NHANES III participants age 50 years and older.," *Diabetes,* p. 1210–1214, 2003.

[19] J. &. R. K. Erikssen, "Resting heart rate in apparently healthy middle-aged men.," *European Journal of Applied Physiology and Occupational Physiology,* pp. 61-69, 1979.

[20] D. T. S. G. E. M. A. R. G. F. C. A. M. Juan Carlos Kaski, "Comparison of epicardial coronary artery tone and reactivity in Prinzmetal's variant angina and chronic stable angina pectoris," *Journal of the American College of Cardiology,* pp. 1058-1062, 1991.

[21] H. G. A. B. E. R. J. K. I. H. A. G. T.A. Kite, "Clinical outcomes of patients discharged from the Rapid Access Chest Pain Clinic with non-anginal chest pain: A retrospective cohort study," *International Journal of Cardiology,* pp. 1-4, 2020.

[22] T. L. M. M. W. S. N. B. K. H. C. Larry W Gibbons, "Maximal exercise test as a predictor of risk for mortality from coronary heart disease in asymptomatic men," *The American Journal of Cardiology,* pp. 53-58, 2000.

[23] Abidin, Z. and Campus, T. The Contribution of Feature Selection and Morphological Operation For On-Line Business System's Image Classification. *International Journal of Multimedia and Ubiquitous Engineering*, 10(11), pp.303-314, 2015

[24] K. G. R. K. S. D. A. W. S. B. S. &. K. D. Vayadande, "Heart Disease Prediction using Machine Learning and Deep Learning Algorithms," in *2022 International Conference on Computational Intelligence and Sustainable Engineering Solutions (CISES)*, 2022.

[25] M. &. P. S. Pal, "Prediction of heart diseases using random forest," *Journal of Physics: Conference Series,* vol. 1817, p. 012009, 2021.

[26] S. N. R. D. M. S. &. H. A. Pasha, "Cardiovascular disease prediction using deep learning techniques.," *IOP conference series: materials science and engineering,* vol. 981, p. 022006, 2020.